\documentclass[letterpaper, 10 pt, conference]{class/ieeeconf}  

\IEEEoverridecommandlockouts                              
\usepackage[table]{xcolor}

\usepackage{float}
\usepackage{verbatim} 
\usepackage{graphicx}
\usepackage{amsmath}
\usepackage{amssymb}
\usepackage{latexsym}
\usepackage{url}
\usepackage{cite}
\usepackage{relsize}
\usepackage{multirow}
\usepackage{afterpage}
\usepackage{ifthen}
\usepackage{graphicx}
\usepackage{algpseudocode,algorithm,algorithmicx}
\usepackage{subfigure}
\usepackage{flushend}
\usepackage{epstopdf}
\usepackage{stackengine}
\usepackage{gensymb}
\usepackage[bookmarks=false, linkcolor=blue, urlcolor=blue, citecolor=blue]{hyperref} 
\usepackage{booktabs}
\usepackage{makecell}
\usepackage{hhline}
\usepackage{courier}
\usepackage{lipsum}
\usepackage{copyrightbox} 

\hypersetup{pdfstartview={FitH}}
\title{\LARGE \bf SurgicalGS: Dynamic 3D Gaussian Splatting for \\ Accurate Robotic-Assisted Surgical Scene Reconstruction}

\author{Jialei Chen$^{1}$, Xin Zhang$^{2}$, Mobarakol Islam$^{3}$, Francisco Vasconcelos$^{3}$, Danail Stoyanov$^{3}$, \\Daniel S. Elson$^{1}$,  Baoru Huang$^{1,3,4}$
\thanks{$^1$The Hamlyn Centre for Robotic Surgery, Imperial College London, SW7 2AZ, UK {\tt Baoru.Huang18@imperial.ac.uk}}
\thanks{$^2$Department of Mechanical and Aerospace Engineering, The HongKong University of Science and Technology, Hong Kong}
\thanks{$^3$ Hawkes Institute, University College London, WC1E 6BT, UK}
\thanks{$^4$Department of Computer Science, University of Liverpool, L69 7ZX, UK}
}
\begin{document}

\newtheorem{problem}{Problem}
\newtheorem{lemma}{Lemma}
\newtheorem{theorem}[lemma]{Theorem}
\newtheorem{claim}{Claim}
\newtheorem{corollary}[lemma]{Corollary}
\newtheorem{definition}[lemma]{Definition}
\newtheorem{proposition}[lemma]{Proposition}
\newtheorem{remark}[lemma]{Remark}
\newenvironment{LabeledProof}[1]{\noindent{\it Proof of #1: }}{\qed}

\def\beq#1\eeq{\begin{equation}#1\end{equation}}
\def\bea#1\eea{\begin{align}#1\end{align}}
\def\beg#1\eeg{\begin{gather}#1\end{gather}}
\def\beqs#1\eeqs{\begin{equation*}#1\end{equation*}}
\def\beas#1\eeas{\begin{align*}#1\end{align*}}
\def\begs#1\eegs{\begin{gather*}#1\end{gather*}}

\newcommand{\poly}{\mathrm{poly}}
\newcommand{\eps}{\epsilon}
\newcommand{\e}{\epsilon}
\newcommand{\polylog}{\mathrm{polylog}}
\newcommand{\rob}[1]{\left( #1 \right)} 
\newcommand{\sqb}[1]{\left[ #1 \right]} 
\newcommand{\cub}[1]{\left\{ #1 \right\} } 
\newcommand{\rb}[1]{\left( #1 \right)} 
\newcommand{\abs}[1]{\left| #1 \right|} 
\newcommand{\zo}{\{0, 1\}}
\newcommand{\zonzo}{\zo^n \to \zo}
\newcommand{\zokzo}{\zo^k \to \zo}
\newcommand{\zot}{\{0,1,2\}}
\newcommand{\en}[1]{\marginpar{\textbf{#1}}}
\newcommand{\efn}[1]{\footnote{\textbf{#1}}}
\newcommand{\vecbm}[1]{\boldmath{#1}} 
\newcommand{\uvec}[1]{\hat{\vec{#1}}}
\newcommand{\thv}{\vecbm{\theta}}
\newcommand{\junk}[1]{}
\newcommand{\var}{\mathop{\mathrm{var}}}
\newcommand{\rank}{\mathop{\mathrm{rank}}}
\newcommand{\diag}{\mathop{\mathrm{diag}}}
\newcommand{\tr}{\mathop{\mathrm{tr}}}
\newcommand{\acos}{\mathop{\mathrm{acos}}}
\newcommand{\atantwo}{\mathop{\mathrm{atan2}}}
\newcommand{\SVD}{\mathop{\mathrm{SVD}}}
\newcommand{\quadf}{\mathop{\mathrm{q}}}
\newcommand{\linterp}{\mathop{\mathrm{l}}}
\newcommand{\sgn}{\mathop{\mathrm{sign}}}
\newcommand{\sym}{\mathop{\mathrm{sym}}}
\newcommand{\avg}{\mathop{\mathrm{avg}}}
\newcommand{\mean}{\mathop{\mathrm{mean}}}
\newcommand{\erf}{\mathop{\mathrm{erf}}}
\newcommand{\grad}{\nabla}
\newcommand{\R}{\mathbb{R}}
\newcommand{\defeq}{\triangleq}
\newcommand{\dims}[2]{[#1\!\times\!#2]}
\newcommand{\sdims}[2]{\mathsmaller{#1\!\times\!#2}}
\newcommand{\udims}[3]{#1}
\newcommand{\udimst}[4]{#1}
\newcommand{\com}[1]{\rhd\text{\emph{#1}}}
\newcommand{\ind}{\hspace{1em}}
\newcommand{\argmin}[1]{\underset{#1}{\operatorname{argmin}}}
\newcommand{\floor}[1]{\left\lfloor{#1}\right\rfloor}
\newcommand{\step}[1]{\vspace{0.5em}\noindent{#1}}
\newcommand{\quat}[1]{\ensuremath{\mathring{\mathbf{#1}}}}
\newcommand{\norm}[1]{\left\lVert#1\right\rVert}
\newcommand{\ignore}[1]{}
\newcommand{\specialcell}[2][c]{\begin{tabular}[#1]{@{}c@{}}#2\end{tabular}}
\newcommand*\Let[2]{\State #1 $\gets$ #2}
\newcommand{\algorithmicbreak}{\textbf{break}}
\newcommand{\Break}{\State \algorithmicbreak}
\newcommand{\ra}[1]{\renewcommand{\arraystretch}{#1}}

\renewcommand{\vec}[1]{\mathbf{#1}} 

\algdef{S}[FOR]{ForEach}[1]{\algorithmicforeach\ #1\ \algorithmicdo}
\algnewcommand\algorithmicforeach{\textbf{for each}}
\algrenewcommand\algorithmicrequire{\textbf{Require:}}
\algrenewcommand\algorithmicensure{\textbf{Ensure:}}
\algnewcommand\algorithmicinput{\textbf{Input:}}
\algnewcommand\INPUT{\item[\algorithmicinput]}
\algnewcommand\algorithmicoutput{\textbf{Output:}}
\algnewcommand\OUTPUT{\item[\algorithmicoutput]}
\maketitle
\thispagestyle{empty}
\pagestyle{empty}

\begin{abstract}

Accurate 3D reconstruction of dynamic surgical scenes from endoscopic video is essential for robotic-assisted surgery. While recent 3D Gaussian Splatting methods have shown promise in achieving high-quality reconstructions with fast rendering speeds, their use of inverse depth loss functions compresses depth variations. This can lead to a loss of fine geometric details, limiting their ability to capture precise 3D geometry and effectiveness in intraoperative application. To address these challenges, we present SurgicalGS, a dynamic 3D Gaussian Splatting framework specifically designed for surgical scene reconstruction with improved geometric accuracy. Our approach first initialises a Gaussian point cloud using depth priors, employing binary motion masks to identify pixels with significant depth variations and fusing point clouds from depth maps across frames for initialisation. We use the Flexible Deformation Model to represent dynamic scene and introduce a normalised depth regularisation loss along with an unsupervised depth smoothness constraint to ensure more accurate geometric reconstruction. Extensive experiments on two real surgical datasets demonstrate that SurgicalGS achieves state-of-the-art reconstruction quality, especially in terms of accurate geometry, advancing the usability of 3D Gaussian Splatting in robotic-assisted surgery.

\end{abstract}


\section{INTRODUCTION} \label{Sec:Intro}
3D reconstruction of surgical scenes is crucial for robotic-assisted surgery. By creating a 3D model of the observed tissues, it allows for more precise instrument control, enabling a range of downstream applications such as intraoperative navigation~\cite{overley2017navigation,huang2020tracking}, robotic surgery automation~\cite{penza2017envisors,huang2024cathaction,zhang2018self,huang2022simultaneous}, and virtual reality simulation~\cite{chong2022virtual}. Traditional methods estimate depth with stereo matching~\cite{song2017dynamic,zhou2019real} or combine simultaneous localisation and mapping (SLAM) to fuse depth map for surgical scene reconstruction~\cite{song2017dynamic,zhou2019real,zhou2021emdq}. However, these methods either assume that scenes are static or surgical tools are absent, limiting their effectiveness in intraoperative applications.

Recently, Neural Radiance Fields (NeRF)~\cite{mildenhall2021nerf} shows a significant progress in scene reconstruction with implicit representations. This technology leverages volume rendering~\cite{drebin1988volume} to convert 2D images to 3D scenes. However, the original NeRF method suffers from long training times, low rendering speed, and a lack of explicit geometric representation~\cite{chen2024survey}. Although some methods adapt discrete structures, such as planes~\cite{cao2023hexplane, fridovich2023k} and voxel grids~\cite{fang2022fast, shao2023tensor4d}, to reduce training time from days to minutes, rendering speed remains insufficient for practical use in surgical scenarios.

\begin{figure}[t]
    \centering
    \includegraphics[width=\linewidth]{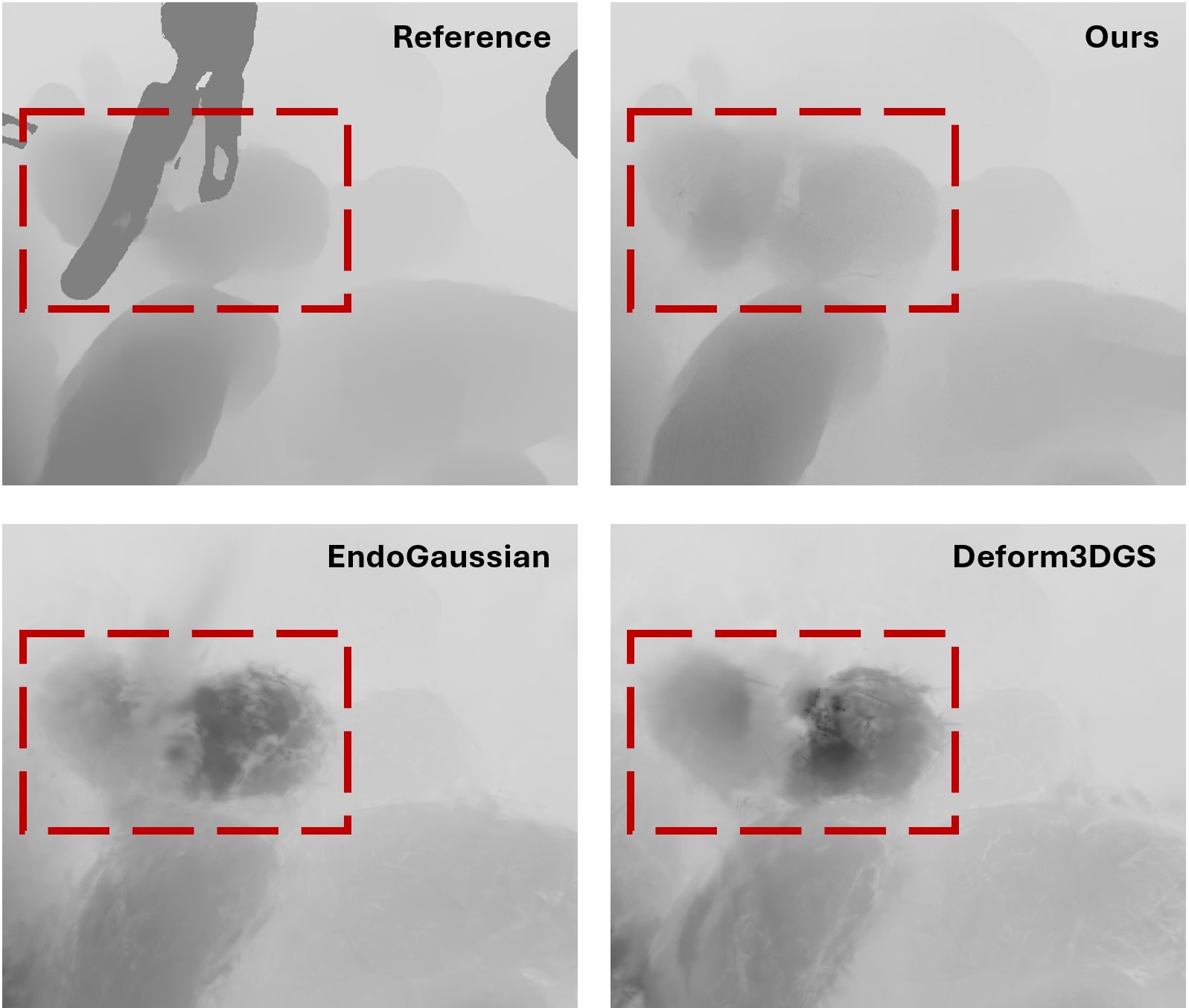}
    \vspace{0.01cm} 
    \caption{We propose a dynamic surgical scene reconstruction frame. Comparisons with EndoGaussian~\cite{liu2024endogaussian} and Deform3DGS~\cite{yang2024deform3dgs} demonstrate that SurgicalGS achieves superior performance in geometric reconstruction.}
    \label{fig:overall}
\end{figure}

To address the limitations of NeRF, 3D Gaussian Splatting (3DGS)~\cite{kerbl20233d} represents the scene as an explicit 3D Gaussian model and renders with an efficient differentiable splatting algorithm~\cite{yifan2019differentiable}, significantly improving the rendering speed to real-time. However, the original 3DGS relies on the Structure-from-Motion (SfM) algorithm~\cite{schonberger2016structure} to initialise Gaussian point clouds, which is not well-suited for surgical scenes with limited viewpoints and frequent tool occlusions. An alternative approach is to use the depth map of the first frame, generated via stereo matching, to initialise the point clouds. However, this method struggles to associate 3D points across frames and fails to reconstruct tissues occluded by tools. Furthermore, 3DGS scenes typically contain millions of Gaussians, whose properties are optimised solely using photometric loss, leading to artefacts and ambiguities due to the lack of 3D cues.

Although previous methods~\cite{liu2024endogaussian,yang2024deform3dgs} have demonstrated satisfactory performance with 3DGS, the application of inverse depth loss functions compresses depth variations. This can lead to a loss of fine geometric details, reducing their ability to capture precise 3D geometry and limiting the effectiveness in intraoperative use. Furthermore, we find that common depth losses, such as L1 and LogL1, often cause memory overflow and instability during training in dynamic scenes. To address these limitations, we focus on two key questions for reconstructing dynamic surgical scenes: (i) How can depth priors be effectively utilised to initialise Gaussian points? (ii) How can depth priors be leveraged to provide accurate depth constraints in dynamic scenes?

In this paper, we propose a new method for accurately reconstructing the dynamic surgical scene, namely SurgicalGS. We first initialise the Gaussian point cloud using depth priors. Specifically, we design a binary motion mask for each frame to identify pixels with significant depth variations. The point clouds extracted from depth maps are fused for initialisation after being downsampled and filtered using motion masks. We then employ a Flexible Deformation Modelling method \cite{yang2024deform3dgs} to represent dynamic scenes. To further enhance depth supervision, we introduce a normalised depth regularisation loss and a depth smoothness constraint to ensure accurate geometric reconstruction. We evaluated our method on two public datasets, \textbf{EndoNeRF}\cite{wang2022neural} and \textbf{StereoMIS}\cite{hayoz2023learning}. Extensive experiments show that SurgicalGS outperforms recent approaches by a large margin and achieves state-of-the-art performance on 3D surgical scene reconstruction. 

Our main contributions and findings are:
\begin{itemize}
    \item We integrate geometric information from all the frames to provide a dense Gaussian initialisation to enhance reconstruction quality.
    \item We propose normalised depth regularisation and an unsupervised depth smoother to incorporate depth priors for better geometry reconstruction.
    \item With extensive experiments we find that (i) SurgicalGS achieves state-of-the-art (SOTA) reconstruction quality, especially with accurate geometry (ii) incomplete and sparse initialisation can lead to a decline in reconstruction performance (iii) strict constraint on the position of the Gaussians can  reduce image quality in dynamic scenes. 
\end{itemize}
\section{Related Work} \label{Sec:rw}
\subsection{Stereo-Matching for MIS}

In robotic-assisted surgery, most endoscopes are equipped with stereo cameras, allowing for depth estimation by stereo matching~\cite{huang2022self}. Previous methods~\cite{song2017dynamic,zhou2019real,zhou2021emdq} combined simultaneous localisation and mapping (SLAM) with stereo depth estimation to fuse depth maps for surgical scene reconstruction. Follow-up approaches~\cite{brandao2021hapnet,huang2021self,luo2022unsupervised,cheng2022deep} leveraged deep learning architectures to enhance the efficiency and accuracy of stereo matching. However, these methods assumed static scenes or the absence of surgical tools, limiting their applicability in real-world scenarios. SuPer \cite{li2020super} and E-DSSR \cite{long2021dssr} addressed these limitations by integrating stereo depth estimation, tool masks, and SurfelWarp \cite{gao2019surfelwarp} to reconstruct surgical scenes from a single viewpoint. Despite these advancements, both methods relied on a sparse warp field \cite{newcombe2015dynamicfusion} to track deformable tissues, which struggled with large non-topological changes and colour alterations resulting from tissue deformation.

\subsection{NeRF-based in Surgery}
Neural Radiance Fields~\cite{mildenhall2021nerf}, an implicit representation using Multi-Layer Perceptrons, demonstrated significant advantages in modelling high-quality appearance. EndoNeRF~\cite{wang2022neural} was the first method to apply NeRF for dynamic surgical scene reconstruction, utilising a canonical field and displacement field to model tissue deformation and canonical density. EndoSurf~\cite{zha2023endosurf} further improved this by incorporating a signed distance function to explicitly constrain the neural field, enabling smoother tissue surface reconstruction. Lerplane~\cite{yang2023neural} encoded temporal and spatial information in 4D volumes and decomposes them into explicit 2D planes to accelerate reconstruction. However, because NeRF relied on dense sampling across millions of rays for implicit representation, these methods suffered from a long training time and low rendering speed, significantly limiting their applicability in intraoperative settings. Achieving high-quality reconstruction with real-time rendering remains a critical challenge.

\subsection{3DGS Applications in Surgery}
Recently, 3D Gaussian Splatting \cite{kerbl20233d} proposed to use anisotropic 3D Gaussians to represent the scene and adopted a differentiable rasterization pipeline to render images, which significantly improves training and rendering speed. To model dynamic scenes with 3DGS, methods like EndoGaussian~\cite{liu2024endogaussian} and work in~\cite{huang2024endo,zhan2024tracking} utilised a deformation field combined with efficient voxel encoding and a lightweight deformation decoder, enabling faster training and rendering for Gaussian tracking. Endo-4DGS~\cite{huang2024endo} adopted monocular depth generated by Depth-Anything~\cite{yang2024depth} as a depth prior and introduced a confidence-guided learning approach to reduce the uncertain measurements in the monocular depth estimation. Deform3DGS~\cite{yang2024deform3dgs} proposed a deformation modelling scheme to represent tissue deformations, which leveraged learnable basis functions in linear combination regression to improve representation capacity and efficiency. Despite achieving high-quality and fast rendering, these methods struggled to reconstruct accurate depth maps, limiting their applicability in intraoperative settings where precise geometry is critical.

\section{Methodology} \label{Sec:method}
\begin{figure*}[h]
    \centering
    \includegraphics[width=\textwidth]{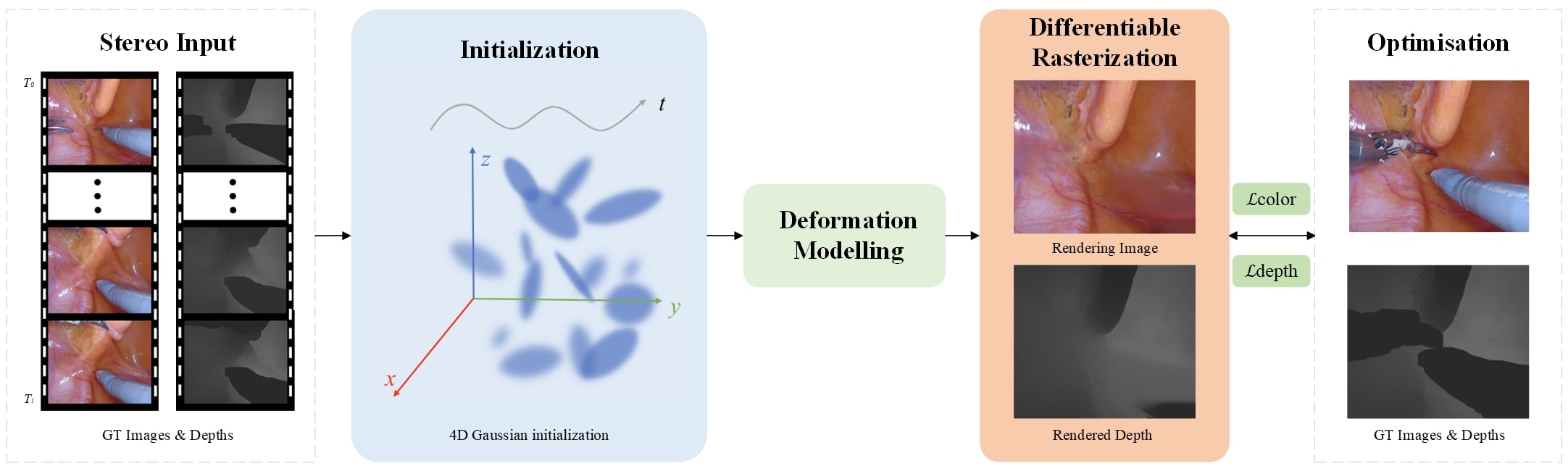}
    \vspace{0.01cm}
    \caption{\textbf{Overall of the proposed framework.} Starting from an image and depth sequence, we associate 3D points between frames to initialise the Gaussian point cloud. Then, given a specific time $t$, the attributes of the Gaussians are modified by the deformation model and rendered as a 2D image and depth map. Finally, colour and depth loss are employed to refine results.}
    \label{fig:pipeline}
\end{figure*}

\subsection{Preliminaries}
\subsubsection{3D Gaussian Splatting}
3DGS~\cite{kerbl20233d} uses the differentiable Gaussians representation to model static scenes, allowing for rapid rasterization and fast image rendering. Each 3D Gaussian consists of learnable attributes: position $\mu$, rotation $r$, scaling $s$, opacity $o$, and spherical harmonic (SH) coefficients. For any 3D point $x$ in the world coordinate, the impact of a 3D Gaussian on $x$ is defined by the Gaussian distribution: 
\begin{equation}
    G(x) = \exp\left( -\frac{1}{2} (x - \mathbf{\mu})^\mathrm{\textit{T}} \Sigma^{-1} (x - \mathbf{\mu}) \right),
    \label{eq:gaussian_pdf}
\end{equation}
\begin{equation}
    \Sigma = R S S^\mathrm{\textit{T}}  R^\mathrm{\textit{T}},
    \label{eq:covariance_decomposition}
\end{equation}
where $\Sigma$ is the covariance matrix, $R$ is a rotation matrix, and $S$ is a diagonal scaling matrix. Then, the 3D Gaussian is projected onto the 2D image planes for rendering. The covariance matrix after projection is calculated by $\Sigma' = J W \Sigma W^T J^T$, where $J$ is the Jacobian of the affine approximation of the projective transformation and $W$ represents the viewing transformation matrix. The final rendering equation for the colour $\hat{C}(x)$ and the depth $\hat{D}(x)$ of the pixel $x$ is: 
\begin{equation}
    \hat{C}(x) = \sum_{i \in N} c_i \alpha_i \prod_{j=1}^{i-1} (1 - \alpha_i), \hat{D}(x) = \sum_{i \in N} d_i \alpha_i \prod_{j=1}^{i-1} (1 - \alpha_i),
    \label{eq:placeholder}
\end{equation}
\begin{equation}
    \alpha_i =  o_iG_i^{2D}(x),
\end{equation}
where $c_i$ is the colour defined from the SH coefficients, $d_i$ is the z-axis in the camera coordinate, and $\alpha_i$ is the density calculated by multiplying the projected 3D Gaussian with the opacity $o_i$. 

\subsubsection{Dynamic Representation}
The original 3DGS~\cite{kerbl20233d} is designed for static scenes. Following \cite{yang2024deform3dgs}, we further represent the dynamic surgical scene using a deformation field, where Fourier and polynomial basis functions $\tilde{b}(t; \theta, \sigma)$ are utilised to learn the motion curve of each Gaussian: 
\begin{equation}
 \tilde{b}(t; \theta, \sigma) = \exp\left( -\frac{1}{2\sigma^2}(t - \theta)^2 \right),
 \label{eq:placeholder}
\end{equation}
where $t$ represents time, and $\theta$ and $\sigma$ denote the learnable centre and variance, respectively. Each Gaussian is associated with a set of learnable weights $\omega$ and parameters $\Theta^\mu$, $\Theta^r$, and $\Theta^s$. These weights and parameters linearly combine the basis functions to represent deformations of position, rotation, and scale, respectively. Taking the position change along the x-axis for an illustration, the deformation can be expressed with a set of parameters $\Theta^\mu_x = \{\omega^\mu_x, \theta^\mu_x, \sigma^\mu_x\}$ as:
\begin{equation}
\psi^{\mu,x}(t; \Theta^{\mu,x}) = \sum_{j=1}^{B} \omega_{j}^{\mu,x} \tilde{b}(t; \theta_{j}^{\mu,x}, \sigma_{j}^{\mu,x}).
\label{eq:placeholder_label}
\end{equation}
The position in x-axis at time t can be expressed as:
\begin{equation}
    \mu_x(t) = \mu_x(t) + \psi^{\mu,x}(t).
\end{equation}

\begin{table*}[h]
\centering
\caption{Quantitative evaluation of our SurgicalGS method against existing methods in endoscopic scene reconstruction. `Speed' denotes the rendering speed (FPS). The optimal and suboptimal results are shown in \textbf{bold} and \underline{underlined} respectively. The unit of depth metrics is millimetre.}
\vspace{0.2cm}
\begin{tabular}{c|c|cccc|ccc|c}
\toprule
\textbf{Dataset} & \textbf{Method} &    \textbf{ABS REL} $\downarrow$&\textbf{SQ REL} $\downarrow$&\textbf{RMSE} $\downarrow$ &\textbf{RMSE LOG} $\downarrow$&\textbf{PSNR} $\uparrow$& \textbf{SSIM} $\uparrow$& \textbf{LPIPS} $\downarrow$ & \textbf{Speed} $\uparrow$\\
\midrule
\multirow{5}{*}{EndoNeRF}& EndoNeRF \cite{wang2022neural} &    0.0232&0.4999& 3.229&0.0234&35.63& 0.941& 0.153& 0.04\\
 & EndoSurf\cite{zha2023endosurf}&   \underline{0.0226}&0.5896&3.075&\textbf{0.0215}& 34.91& 0.953& 0.112&0.04\\
 & LerPlane\cite{yang2023neural}& 0.0351& 0.5717& 5.734& 0.0275& 35.00& 0.927& 0.099&0.93\\
& EndoGaussian\cite{liu2024endogaussian} &    0.0340&\underline{0.2362}&\underline{2.926}&0.0343&37.71& 0.958& 0.062& 148.35\\
 & Deform3DGS\cite{yang2024deform3dgs}&   0.0324&0.2810&3.275&0.0318& \textbf{38.39}& \textbf{0.962}& \textbf{0.059}&\textbf{332.52}\\
\rowcolor{gray!30} \cellcolor{white}& SurgicalGS (Ours) &    \textbf{0.0219}&\textbf{0.1115}& \textbf{1.820}&\underline{0.0219}&\underline{38.18}& \underline{0.960}& \underline{0.062}& \underline{194.80}\\
\midrule
\multirow{5}{*}{StereoMIS}& EndoNeRF\cite{wang2022neural} &    0.0315&0.7183& 3.022&0.0191&28.79& 0.809& 0.266& 0.06\\
 & EndoSurf\cite{zha2023endosurf}&   \underline{0.0189}&\underline{0.4031}& \underline{2.457}&\underline{0.0123}& 29.36& 0.861& 0.211&0.05\\
 & LerPlane\cite{yang2023neural}& 0.0354& 1.0906& 5.521& 0.0362& 29.09& 0.789& 0.179&0.95\\
& EndoGaussian\cite{liu2024endogaussian}&    0.0292&0.6057&5.050&0.0308&31.02& 0.878& \textbf{0.132}& 130.15\\
 & Deform3DGS\cite{yang2024deform3dgs}&   0.0330&0.7547&4.888&0.0330& \textbf{31.61}& \textbf{0.888}& \underline{0.135}&\textbf{308.66}\\
\rowcolor{gray!30} \cellcolor{white}& SurgicalGS (Ours) &    \textbf{0.0082}&\textbf{0.0391}&\textbf{2.174}&\textbf{0.0082}&\underline{31.54}& \underline{0.885}& 0.148& \underline{214.06}\\
\bottomrule
\end{tabular}

\label{tab:comparison}
\end{table*}

\subsection{Proposed Method: SurgicalGS}
Our proposed SurgicalGS integrates geometric information from all frames for dense Gaussian initialisation to recover tissues occluded by tools and improve reconstruction quality. Additionally, it employs normalised depth loss and unsupervised depth smoothness to enhance the accuracy of geometric reconstruction.
\subsubsection{Dense Initialisation with Depth Priors}
The original 3DGS \cite{kerbl20233d} used the SfM algorithm \cite{schonberger2016structure} to generate initialised point clouds for further reconstruction. However, due to limited viewpoints, dynamic lighting conditions, and tool occlusions in medical environments, SfM cannot generate accurate point clouds in surgical scenes. Thus, we employ the depth map, tissue mask, and camera parameters to extract point clouds of tissues for each frame as:
\begin{equation}
P_i = K_1^{-1}K_2^{-1} D_i (I_i \odot M_i),
\end{equation}
where $P_i$ denotes the 3D point cloud of the $i$-th frame, and $D_i$, $I_i$, and $M_i$ represent the depth map from stereo-matching, input image, and binary tissue mask for the $i$-th frame, respectively, $K_1$ and $K_2$ refer to the known camera intrinsic and extrinsic matrices, respectively, and $\odot$ denotes element-wise multiplication. However, the point cloud is incomplete due to surgical tool occlusions. We observe that the occluded tissue in the $i$-th frame could be visible in some of the other frames. Furthermore, a dense initialisation of the Gaussian point cloud can more accurately represent the scene \cite{kerbl20233d}, leading to improved reconstruction quality for deformable tissues. Based on these observations, we design a binary motion mask $B_i$ for each frame to extract pixels with significant depth variance and occluded tissues in the first frame:
\begin{equation}
B_i = \mathbb{I}\left( \left| D_0 - D_i \right| > \tau \right) \cup (1 - M_0) \cap M_i,
\end{equation}
where $\mathbb{I}(\cdot)$ refers to the indicator function, $M_0$ denotes the tissue mask of the first frame, and $\tau$ is the threshold defines the significant depth variance. We uniformly downsample the point clouds to reduce the number of points and fuse them to initialise the Gaussian point cloud as follow:
\begin{equation}
P = \{P_0, P_1\odot B_1, ..., P_T \odot B_T\},
\end{equation}
where $T$ refers to the frame length.

\subsubsection{Normalised Depth Regularisation}
To ensure accurate alignment between the predicted and actual depth maps, depth regularisation is applied to supervise the 3D reconstruction learning process. A common approach is to use $\mathcal{L}_{\text{1}}$ depth loss. However, in dynamic scenes, we observe that directly applying $\mathcal{L}_{\text{1}}$ depth loss results in an overly dense Gaussian point cloud, causing memory overflow and instability in training.

Previous methods \cite{yang2024deform3dgs, liu2024endogaussian} incorporate inverse depth maps into the loss computation, effectively stabilising the optimisation process. The depth loss in these approaches is formulated as:
\begin{equation}
    \mathcal{L}_{\hat{\text{D}}}^{-1} = \| M \odot (\hat{D}^{-1} - D^{-1}) \|,
    \label{eq:depth_loss}
\end{equation}
where $D$ and $\hat{D}$ are the depth map from stereo-matching and rendered depth, respectively. Inverse depth maps compress the dynamic range of depth values, reducing the disparity between the binocular and rendered depth maps. This compression minimises the risk of over-density and enhances the stability of the optimisation process. However, there is little variation in the depth map of endoscopic videos. Using inverse depth maps can overly homogenise the depth values, resulting in inaccurate and inconsistent rendered depth maps.

To address this problem, our observation is that normalisation can bring both binocular and rendered depth maps in a consistent scale, ensuring training stability while preserving depth variability. Our normalised depth loss is formulated as:
\begin{equation}
    \mathcal{L}_{\hat{\text{D}}} = \| M \odot (\hat{D}_{\text{norm}} - D_{\text{norm}}) \|.
    \label{eq:depth_loss}
\end{equation}

\subsubsection{Unsupervised Depth Smoothness}
In surgical scenes, specular highlights, homogeneous surfaces, and large disparity discontinuities make it difficult for stereo matching algorithms to establish accurate correspondences, leading to noise in the depth maps \cite{stoyanov2012surgical}. To remove the influence of noise and enforce the smoothness of the rendered depth, we employ a total variation loss. In addition, to preserve depth details, we apply the Canny edge detector \cite{canny1986computational} as a mask to prevent the regularisation of edges with significant depth variations. We regularise the difference between a pixel $\hat{D}_{i,j}$ in depth map and its adjacent pixel as:
\begin{align}
    \mathcal{L}_{\text{smooth}} = \frac{1}{|{\hat{D}}|} \sum_{i,j} \mathbb{I}_{\text{ne}}(\hat{D}_{i,j}) \cdot & \left( |\hat{D}_{i,j} - \hat{D}_{i+1,j}| \right. \notag \\
    & \left. + |\hat{D}_{i,j} - \hat{D}_{i,j+1}| \right)
    \label{eq:smooth_loss}
\end{align}
where $\mathbb{I}_{\text{ne}}(\hat{D}_{i,j})$ is the result of Canny edge detection indicating whether the pixel $(i,j)$ is on the edge.

\subsection{Total Loss Function}
The final loss for optimisation is defined as:
\begin{equation}
    \mathcal{L} = \mathcal{L}_{\text{colour}} + 
    \underbrace{\left( \mathcal{L}_{\hat{\text{D}}} + \lambda_{\text{smooth}} \mathcal{L}_ {\text{smooth}}\right)}_{\mathcal{L}_{\text{depth}}}
    \label{placeholder_label}
\end{equation}
where $\mathcal{L}_{\text{colour}}$ is the original photometric loss in 3DGS \cite{kerbl20233d}.

\section{Experiments and Results} \label{Sec:exp}

\subsection{Experiment Setting}
\textbf{Dataset and Evaluation Metrics.} We evaluate the proposed method and compare it with existing methods on two public datasets: (i) the EndoNeRF dataset~\cite{wang2022neural}, which contains \textit{in-vivo} prostatectomy data captured from stereo cameras at a single viewpoint and provides estimated depth maps with stereo matching~\cite{li2021revisiting} and manually labelled tool masks. (ii) the StereoMIS dataset~\cite{hayoz2023learning}, which is a stereo video dataset captured from \textit{in-vivo} porcine subjects containing large tissue deformations. Tool masks are also provided. We estimate depth maps using the pre-trained RAFT model~\cite{teed2020raft}. Following~\cite{zha2023endosurf}, we divide frames of each scene into training and testing sets with a 7:1 ratio. We use PSNR, SSIM, and LPIPS to evaluate the similarity between actual and rendered images, common depth metrics, similar to~\cite{kusupati2020normal, luo2020consistent, murez2020atlas, sinha2020deltas}, to measure the quality of the depth map, and frame per second (FPS) to evaluate reconstruction efficiency.

\begin{figure*}[t]
    \centering
    \includegraphics[width = 0.9\textwidth,height=0.9\linewidth]{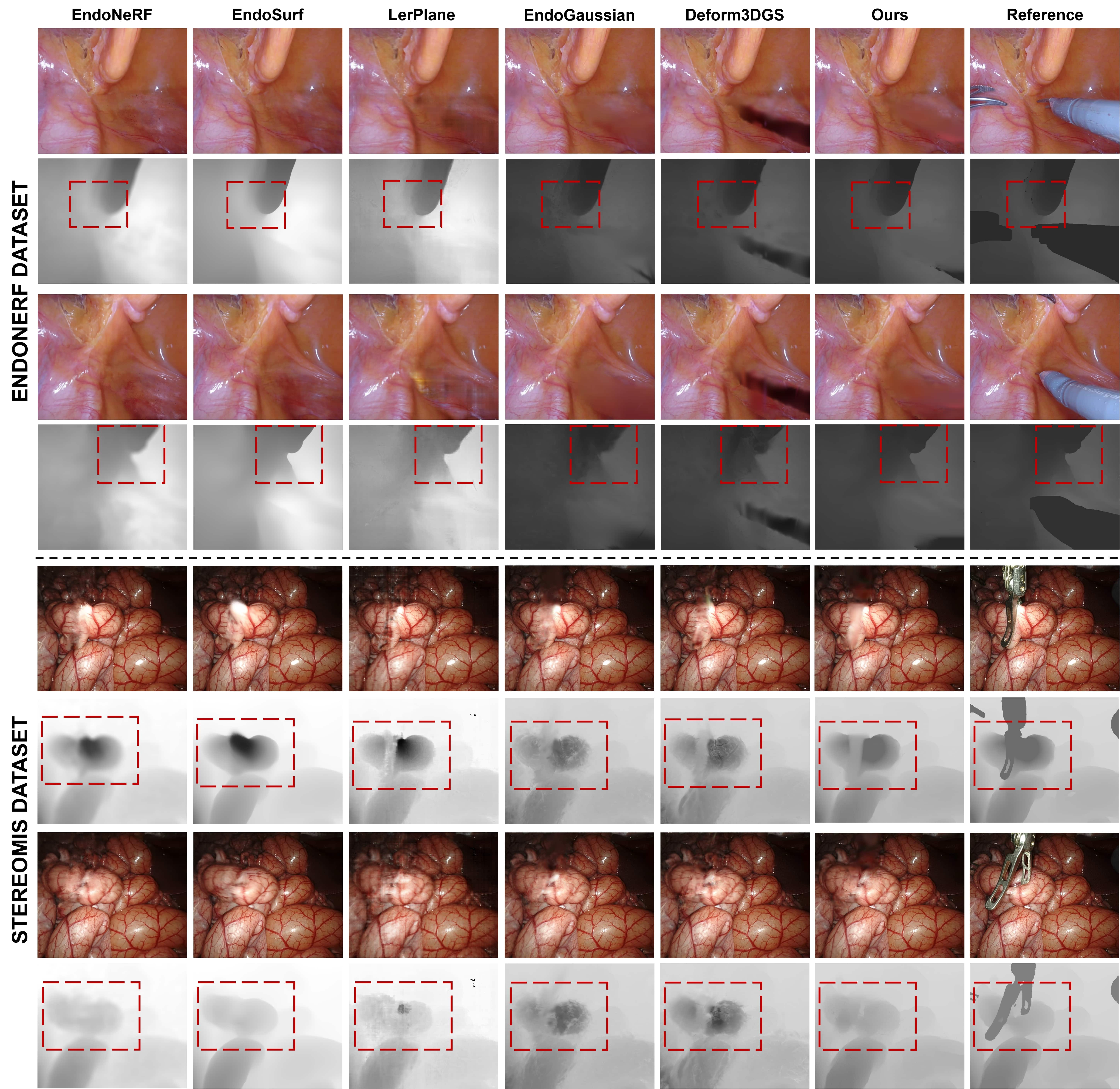}
    \vspace{0.3cm} 
    \caption{Visualisation of the 3D reconstruction results. Further details can be seen by zooming in.}
    \label{fig:ENDONERF_result}
\end{figure*}

\textbf{Implementation Details.} All the experiments are conducted on the RTX3090 GPU and the PyTorch framework. We employ the Adam optimiser with an initial learning rate of $1.6 \times 10^{-3}$. For all scenes, the model is trained for 6K iterations with the same loss function and initialisation strategy. We set $\lambda_{\text{smooth}} = 0.0001$ in our experiments.

\begin{table}[t] 
\centering
\caption{Ablation of depth loss on StereoMIS dataset. $\mathcal{L}_{\text{LogL1}}$ is logarithmic L1 depth loss, $\mathcal{L}_{\hat{\text{D}}}^{-1}$ is inverse depth loss, $\mathcal{L_{\hat{\text{D}}}}$ is our normalised depth loss, and $\mathcal{L}_{\text{depth}}$ is normalised depth loss along with unsupervised depth smoothness.}
\vspace{0.2cm}
\resizebox{\linewidth}{!}{ 
\begin{tabular}{l|ccc}
 \toprule
 \textbf{Method}&\textbf{RMSE} $\downarrow$ &\textbf{PSNR} $\uparrow$& \textbf{SSIM(\%)} $\uparrow$ \\
 \midrule
 ours+$\mathcal{L}_{\text{LogL1}}$& 2.375&23.44& 0.643\\
  ours+$\mathcal{L}_{\text{1}}$&2.436& 22.91& 0.630\\
 ours+$\mathcal{L}_{\hat{\text{D}}}^{-1}$&4.956&\textbf{32.29}& \textbf{0.903}\\
 \midrule
 ours+$\mathcal{L}_{\hat{\text{D}}}$& \underline{2.256}&31.54& 0.885\\
 ours+$\mathcal{L}_{\text{depth}}$& \textbf{2.174}& \underline{31.54}& \underline{0.885}\\
 \bottomrule
\end{tabular}%
}
\label{tab:ablation_depth}
\end{table}

\begin{figure}[h]
    \centering
    \includegraphics[width=\linewidth]{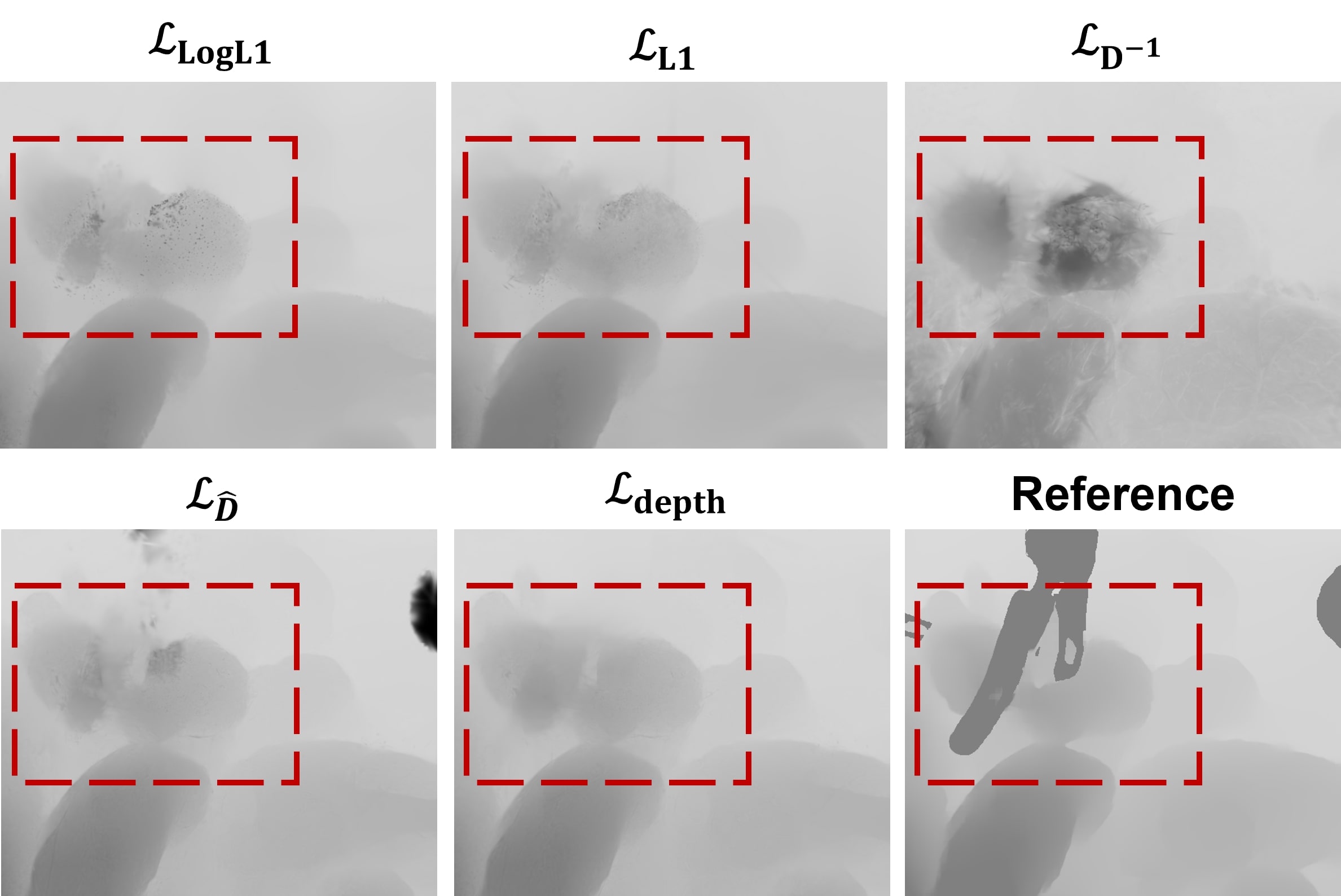}
    \vspace{0.01cm} 
    \caption{Visualisation of ablation study on depth loss using the StereoMIS dataset. Further details can be seen by zooming in.}
    \label{fig:ablation}
\end{figure}

\begin{table}[t]
\centering
\caption{Ablation of initialisation strategy on StereoMIS dataset. `w/o Init' denotes initialisation with the depth map of the first frame.}
\vspace{0.2cm} 
\resizebox{\linewidth}{!}{ 
\begin{tabular}{c|ccc}
\toprule
 \textbf{Method} &\textbf{RMSE} $\downarrow$ &\textbf{PSNR} $\uparrow$& \textbf{SSIM(\%)} $\uparrow$ \\
\midrule
 w/o Init& 2.362&31.43& 0.881\\
  w/ Init&\textbf{2.174}& \textbf{31.54}& \textbf{0.885}\\
\bottomrule
\end{tabular}
}
\label{tab:init_ablation}
\end{table}

\subsection{Qualitative and Quantitative Evaluation}
We evaluate our method by comparing its performance with EndoNeRF \cite{wang2022neural}, LerPlane~\cite{yang2023neural}, the SOTA on surface reconstruction: EndoSurf \cite{zha2023endosurf}, and two SOTAs on fast and high-quality reconstruction: EndoGaussian \cite{liu2024endogaussian} and Deform3DGS \cite{yang2024deform3dgs}. As listed in Table~\ref{tab:comparison}, although EndoNeRF\cite{wang2022neural}, EndoSurf\cite{zha2023endosurf}, and LerPlane~\cite{yang2023neural} effectively reconstruct deformable tissues, they suffer from low rendering speed and struggle with rendering high-quality images, which limits their effectiveness for real-time surgical scene reconstruction. On the other hand, EndoGaussian \cite{liu2024endogaussian} and Deform3DGS \cite{yang2024deform3dgs} improve image quality and rendering speed based on 3DGS, but they fail to reconstruct accurate depth maps, limiting their intraoperative reliability. Benefiting from normalised depth regularisation, our method achieves superior performance in accurate geometric reconstruction while maintaining a suboptimal result in rendering quality and speed. Although the rendering speed of our method is slower than that of Deform3DGS\cite{yang2024deform3dgs}, it still achieves real-time performance, as most real-world endoscopes operate at 30-60 FPS.

As shown in Fig. \ref{fig:ENDONERF_result}, we also visualise several scenes for qualitative evaluation. It can be seen that our method effectively models complex tissue motions and preserves texture details. In addition, our method reconstructs an accurate and smooth depth map. In contrast, EndoNeRF~\cite{wang2022neural}, EndoSurf~\cite{zha2023endosurf}, and LerPlane~\cite{yang2023neural} fail to adjust scaling to match the real depth distribution, while EndoGaussian~\cite{liu2024endogaussian} and Deform3DGS~\cite{yang2024deform3dgs} struggle with capturing precise edges, especially in regions with deformable tissues. These results demonstrate that our method achieves accurate reconstructions of the dynamic surgical scene, highlighting its potential for applications in intraoperative navigation and robotic surgery automation.

\subsection{Ablations and Analysis}

We first analyse the effect of our proposed normalised depth loss $\mathcal{L_{\hat{\text{D}}}}$ on the StereoMIS dataset. We compare it with common depth losses, such as $\mathcal{L}_{\text{1}}$ and $\mathcal{L}_{\text{LogL1}}$, as well as inverse depth loss $\mathcal{L}_{\hat{\text{D}}}^{-1}$ used in Deform3DGS\cite{yang2024deform3dgs} and EndoGaussian\cite{liu2024endogaussian}. As is shown in Table~\ref{tab:ablation_depth}, $\mathcal{L_{\hat{\text{D}}}}$ generally outperforms other depth losses on depth and colour metrics and the unsupervised depth smoothness with Eq. \ref{eq:smooth_loss} has a slight improvement in depth metrics. Additionally, we find that although $\mathcal{L}_{\hat{\text{D}}}^{-1}$ underperforms on depth metrics, it significantly improves image quality. This may be because it imposes less constraint on the position of the Gaussians, allowing them to represent the scene more freely. Furthermore, in Fig. \ref{fig:ablation}, it can be seen that both $\mathcal{L}_{\text{1}}$ and $\mathcal{L}_{\text{LogL1}}$ introduce noise and speckling artefacts, while $\mathcal{L}_{\hat{\text{D}}}^{-1}$ results in inaccurate geometry comparing with the actual scene. In contrast, our $\mathcal{L_{\hat{\text{D}}}}$ produces a more accurate depth map, and the unsupervised depth smoothness significantly enhances the overall visual quality. We further evaluate the effect of our dense initialisation strategy in Table \ref{tab:init_ablation}. The `w/o init' denotes using point clouds of the first frame to initialise the Gaussian points, which results in a performance decrease, showing the significance of the proposed dense Gaussian initialisation.

\section{Conclusions}\label{Sec:con}

In this paper, we present a novel 3DGS-based approach for accurate robotic-assisted surgical scene reconstruction. Different from previous methods, we proposed a normalised depth regularisation and unsupervised depth smoother to ensure accurate geometric reconstruction. Additionally, the dense initialisation strategy is introduced to improve reconstruction quality. From the ablation study, we validate the effects of our proposed methods. We also observe that L1 and LogL1 depth loss introduce noise and significantly degrade reconstruction quality. Additionally, inverse depth loss struggles to capture accurate geometry, but it improves image quality. This might be due to it imposes less constraint on the position of the Gaussians, allowing them greater freedom to represent the scene. Furthermore, incomplete and sparse initialisation also leads to a noticeable decline in reconstruction performance. Extensive experiments on two real surgical datasets show that our method achieves state-of-the-art reconstruction quality, particularly in terms of geometric accuracy. 

Future work will focus on evaluating the robustness of SurgicalGS under a variety of real-world and synthetic surgical scenarios, including complex and challenging conditions such as blood, smoke, and blurring. Additionally, further exploration is needed to assess the model’s ability to maintain high reconstruction quality and geometric accuracy in these adverse conditions, ensuring its adaptability and reliability in more dynamic and unpredictable surgical environments. Our code and model will be released.

\bibliographystyle{class/IEEEtran}
\bibliography{class/IEEEabrv,class/reference}

\end{document}